\documentclass[10pt,twocolumn,letterpaper]{article}

\usepackage[pagenumbers]{cvpr} 
\usepackage[accsupp]{axessibility}  

\newcommand{\blfootnote}[1]{%
  \begingroup
  \renewcommand\thefootnote{}\footnote{#1}%
  \addtocounter{footnote}{-1}%
  \endgroup
}

\definecolor{cvprblue}{rgb}{0.21,0.49,0.74}
\usepackage[pagebackref,breaklinks,colorlinks,allcolors=cvprblue]{hyperref}

\title{GeomPrompt: Geometric Prompt Learning for RGB-D\\Semantic Segmentation Under Missing and Degraded Depth}

\author{
Krishna Jaganathan,\; Patricio Vela\\
Georgia Institute of Technology\\
{\tt\small \{kjaganathan7, pvela\}@gatech.edu}
}

\begin{document}

\twocolumn[{%
\renewcommand\twocolumn[1][]{#1}%
\maketitle

\vspace{-2em}
\begin{center}
    \includegraphics[width=.9825\textwidth]{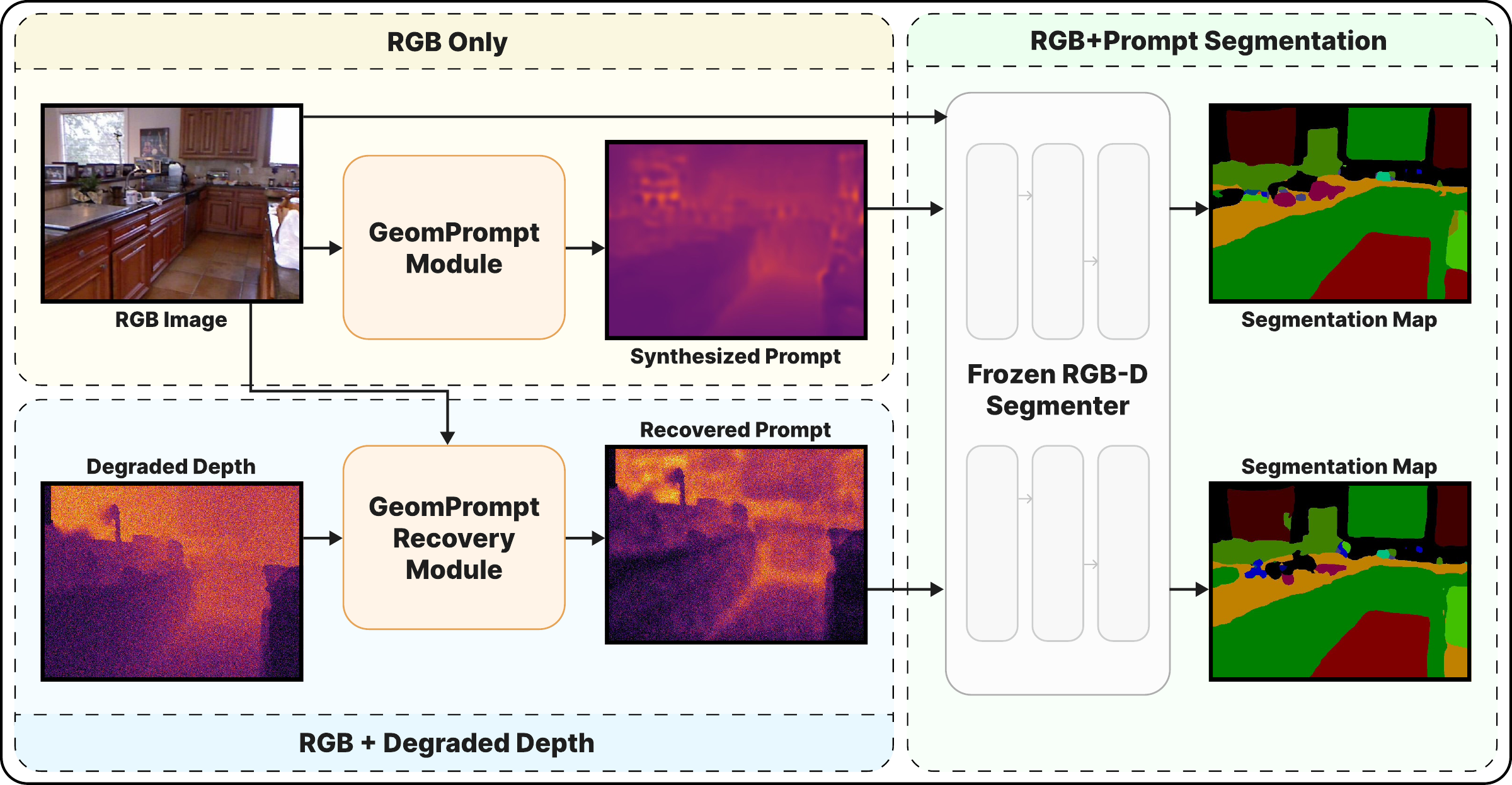}
    \vspace{-0.25em}
    \captionof{figure}{\textit{GeomPrompt} synthesizes a geometric prompt from an RGB image for downstream semantic segmentation on a frozen RGB-D segmenter in the case of missing depth. \textit{GeomPrompt-Recovery} extends this by allowing for corrections on existing degraded depth inputs.}
    \label{fig:teaser}
\end{center}
\vspace{0.5em}
}]

\blfootnote{Supported in part by NSF Award \#2345057.}
\blfootnote{Project page: \url{https://geomprompt.github.io}}

\begin{abstract}
Multimodal perception systems for robotics and embodied AI often assume reliable RGB-D sensing, but in practice, depth is frequently missing, noisy, or corrupted. We thus present GeomPrompt, a lightweight cross-modal adaptation module that synthesizes a task-driven geometric prompt from RGB alone for the fourth channel of a frozen RGB-D semantic segmentation model, without depth supervision. We further introduce GeomPrompt-Recovery, an adaptation module that compensates for degraded depth by predicting the fourth channel correction relevant for the frozen segmenter. Both modules are trained solely with downstream segmentation supervision, enabling recovery of the geometric prior useful for segmentation, rather than estimating depth signals. On SUN RGB-D, GeomPrompt improves over RGB-only inference by +6.1 mIoU on DFormer and +3.0 mIoU on GeminiFusion, while remaining competitive with strong monocular depth estimators. For degraded depth, GeomPrompt-Recovery consistently improves robustness, yielding gains up to +3.6 mIoU under severe depth corruptions. GeomPrompt is also substantially more efficient than monocular depth baselines, reaching 7.8 ms latency versus 38.3 ms and 71.9 ms. These results suggest that task-driven geometric prompting is an efficient mechanism for cross-modal compensation under missing and degraded depth inputs in RGB-D perception.
\end{abstract}    
\section{Introduction}
\label{sec:intro}

Semantic segmentation is a core perception primitive for embodied systems because it converts raw sensor observations into structured scene understanding that supports action. In robotic settings, segmentation provides object-level and layout-level priors that help agents reason about what is present, where it is, and how to move or interact safely in cluttered environments. These semantic cues are useful for downstream tasks such as object-goal navigation, semantic mapping, and mobile manipulation, where success depends not only on recognizing categories but also on grounding those categories in spatially meaningful scene structure~\cite{raychaudhuri_semantic_2025, sun_survey_2025, crespo_semantic_2020}.

A common way to strengthen semantic segmentation is to incorporate geometry, especially through RGB-D sensing. Depth provides complementary structural information that can improve pixel-level prediction beyond what is available from appearance alone, and RGB-D segmentation has thus become a standard bimodal setting in both the segmentation and robotics literature~\cite{wang_brief_2021, yin_dformer:_2023, jia_geminifusion:_2024, yin_dformerv2:_2025}. In this regime, depth is valuable as a prior that can sharpen boundaries, separate objects with similar appearance, and improve scene layout reasoning~\cite{wang_brief_2021, yin_dformerv2:_2025}.

However, this reliance on depth creates a practical mismatch at deployment time. In real robotic systems, depth can be unavailable, spatially incomplete, noisy, quantized, or otherwise unreliable due to sensor failures, difficult materials, environmental conditions, or range and hardware limitations~\cite{liao_benchmarking_2025, haider_what_2022, he_depth_2017, stommel_inpainting_2014}. Reflective and transparent surfaces, for example, are well known to produce corrupted or missing measurements, and even when depth is present, its quality can vary substantially across sensors and operating conditions~\cite{he_depth_2017, huang_polarization_2023, stommel_inpainting_2014}. As a result, the settings in which geometric cues are most helpful for perception are often exactly those in which depth is hardest to trust~\cite{liao_benchmarking_2025}.

Existing alternatives partially resolve this gap. Some approaches use depth or geometry as privileged information during training, while others insert monocular depth estimation as an intermediate step before segmentation~\cite{jiao_geometry-aware_2019, gu_hard_2021, godard_digging_2019, yang_depth_2024, hu_metric3d_2024}. These strategies can be effective, but they typically still depend on explicit geometric supervision, depth-oriented pretraining, synthetic RGB-D data, pseudo-depth pipelines, or additional estimation modules whose objective is to reconstruct metric or proxy geometry rather than directly optimize the downstream segmentation task~\cite{yang_depth_2024, hu_metric3d_2024, khan_deep_2020}. This leaves open a simpler question: When a strong RGB-D segmenter expects geometric input, can we learn to supply only a geometry-like signal sufficient for the segmenter, without supervising that signal as depth at all~\cite{jiao_geometry-aware_2019, gu_hard_2021}?

We address this question with \textit{GeomPrompt}, a lightweight module that learns to synthesize a geometric prompt from RGB for a frozen RGB-D segmenter. Rather than reconstructing metric depth, GeomPrompt learns the geometry-like signal that is useful for the downstream multimodal model, through segmentation-only supervision. We further introduce \textit{GeomPrompt-Recovery}, a conditioned variant for the complementary setting in which depth is present but unreliable. Given RGB together with degraded depth, it predicts a bounded residual correction that restores task-relevant geometric information for the frozen segmenter under segmentation-only supervision. In this view, GeomPrompt and GeomPrompt-Recovery act as lightweight mechanisms for cross-modal guidance and compensation in RGB-D perception under sensor failure.  

This perspective reframes the problem from estimating depth to recovering the geometric prior relevant for RGB-D segmentation under sensor failure. Empirically, our results show that this simple prompting view yields a resource-aware alternative to explicit monocular depth estimation, as it preserves frozen RGB-D backbones, improves performance under missing and degraded depth, and remains lightweight enough to be attractive for resource-constrained embodied perception.

Our core contributions are as follows.
\begin{enumerate}
    \item \textit{GeomPrompt}, a lightweight cross-modal adaptation module for frozen RGB-D segmentation under missing depth, which synthesizes a task-relevant geometric prompt from RGB alone.
    \item \textit{GeomPrompt-Recovery}, a degradation-aware recovery module that predicts task-relevant corrections for degraded depth rather than reconstructing metric depth.
    \item Evidence that segmentation-only supervision can support robust multimodal segmentation under sensor unreliability, improving missing and degraded modality performance while remaining efficient.
\end{enumerate}

\section{Related Work}
\label{sec:relatedwork}
\subsection{RGB-D Semantic Segmentation}
Recent RGB-D semantic segmentation methods improve performance by fusing depth with RGB through dual-stream encoders, cross-modal attention, and geometry-aware priors. Representative approaches such as DFormer~\cite{yin_dformer:_2023}, GeminiFusion~\cite{jia_geminifusion:_2024}, CMX~\cite{zhang_cmx:_2023}, DFormerv2~\cite{yin_dformerv2:_2025}, and AsymFormer~\cite{du_asymformer:_2024} span efficient multimodal fusion, explicit geometric priors, and real-time asymmetric designs, and together establish strong baselines for RGB-X segmentation.

A broader family of works further explores scale-invariant depth encoding, attention-based fusion, and efficient backbone design across RGB-D architectures~\cite{wan_sigma:_2025,wang_multimodal_2022,wang_deep_2020,seichter_efficient_2022,woo_real-time_2025,chen_depth_2025,zhong_attention-based_2024,wu_transformer_2024,lu_fcegnet:_2025,xu_interactive_2023}. Despite their strong empirical performance, these methods are typically designed under the assumption that depth is available and sufficiently reliable at inference time.

\subsection{Geometry as Privileged Information}
To reduce reliance on test-time depth, several approaches use geometry as privileged information available only during training. Methods such as Geometry-Aware Distillation~\cite{jiao_geometry-aware_2019} and Hard Pixel Mining for Depth Privileged Semantic Segmentation~\cite{gu_hard_2021} use distillation or depth privileged supervision to improve RGB segmentation without requiring depth at inference.

Related works also leverage depth or synthetic geometric structure for domain adaptation and unsupervised representation learning~\cite{lee_spigan:_2018,vu_dada:_2019,chen_learning_2019,sick_unsupervised_2024}. In contrast, GeomPrompt does not predict, regress, reconstruct, or distill depth explicitly. Instead, it learns a task-relevant auxiliary prompt for a frozen RGB-D segmenter using segmentation loss alone.

\subsection{Estimated Depth as Proxy Geometry}
Another common strategy is to synthesize or estimate proxy depth to compensate for missing or unreliable sensor measurements. This includes modern supervised and metric monocular depth estimators such as Depth Anything V2~\cite{yang_depth_2024}, Metric3Dv2~\cite{hu_metric3d_2024}, and UniDepthV2~\cite{piccinelli_unidepthv2:_2026}, as well as self-supervised approaches such as Monodepth2~\cite{godard_digging_2019}. Many segmentation pipelines follow a two-stage paradigm in which depth is first estimated and then consumed by the segmentation model~\cite{cardace_plugging_2022,guo_semantic_2018}, while others use depth-oriented pretraining before segmentation fine-tuning~\cite{leonardis_viability_2025}.

Additional methods introduce pseudo-depth or estimated depth assistance through diffusion models or semi-supervised auxiliary training~\cite{xu_pddm:_2025, sun_semantic_2021, lagos_semsegdepth:_2022, hoyer_improving_2023}. By contrast, GeomPrompt does not rely on explicit depth prediction, depth-oriented pretraining, real or synthetic depth supervision, or photometric self-supervision.

\subsection{Missing Modality Robustness}
A growing body of literature studies semantic segmentation under missing modalities or degraded sensor inputs. Techniques such as Depth Removal Distillation~\cite{fang_depth_2022} explicitly reduce dependence on depth, while methods such as M3L~\cite{maheshwari_missing_2024}, MAGIC~\cite{leonardis_centering_2025}, and Delivering Arbitrary-Modal Semantic Segmentation~\cite{zhang_delivering_2023} train multimodal architectures to operate under missing or varying modality availability. Recent benchmarks further highlight the fragility of multimodal segmentation systems under realistic sensor failures and modality corruption~\cite{liao_benchmarking_2025}.

Related robustness architectures address fusion robust to noise, along with degraded depth handling and diffusion-based refinement within RGB-D segmentation pipelines~\cite{woo_real-time_2025,chen_depth_2025,zhong_attention-based_2024,bui_diffusion-based_2025}. In contrast to methods that train or adapt the multimodal backbone itself for robustness, both GeomPrompt and GeomPrompt-Recovery act as lightweight adaptation modules for a frozen RGB-D segmenter under missing (GeomPrompt) or corrupted/quantized (GeomPrompt-Recovery) depth.

\section{Methodology}
\label{sec:methodology}

\subsection{Prompting Framework}
We consider a frozen RGB-D segmenter $S$ deployed under two conditions, namely missing depth, where only RGB $x$ is available, and degraded depth, where RGB is paired with corrupted depth $\tilde d$. Rather than reconstructing metric depth, we learn a task-driven geometric prompt $p^*$ in the depth input space expected by $S$. In the missing depth setting, GeomPrompt predicts $p^*=G(x)$, and in the degraded depth setting, GeomPrompt-Recovery predicts $p^*=G_R(x,\tilde d)$. The final segmentation prediction is given by $\hat y=S(x,p^*)$. Both modules are trained only with segmentation supervision from $y$, without depth supervision.

\begin{figure*}
  \centering
  \begin{subfigure}{0.61\linewidth}
    \includegraphics[width=\linewidth]{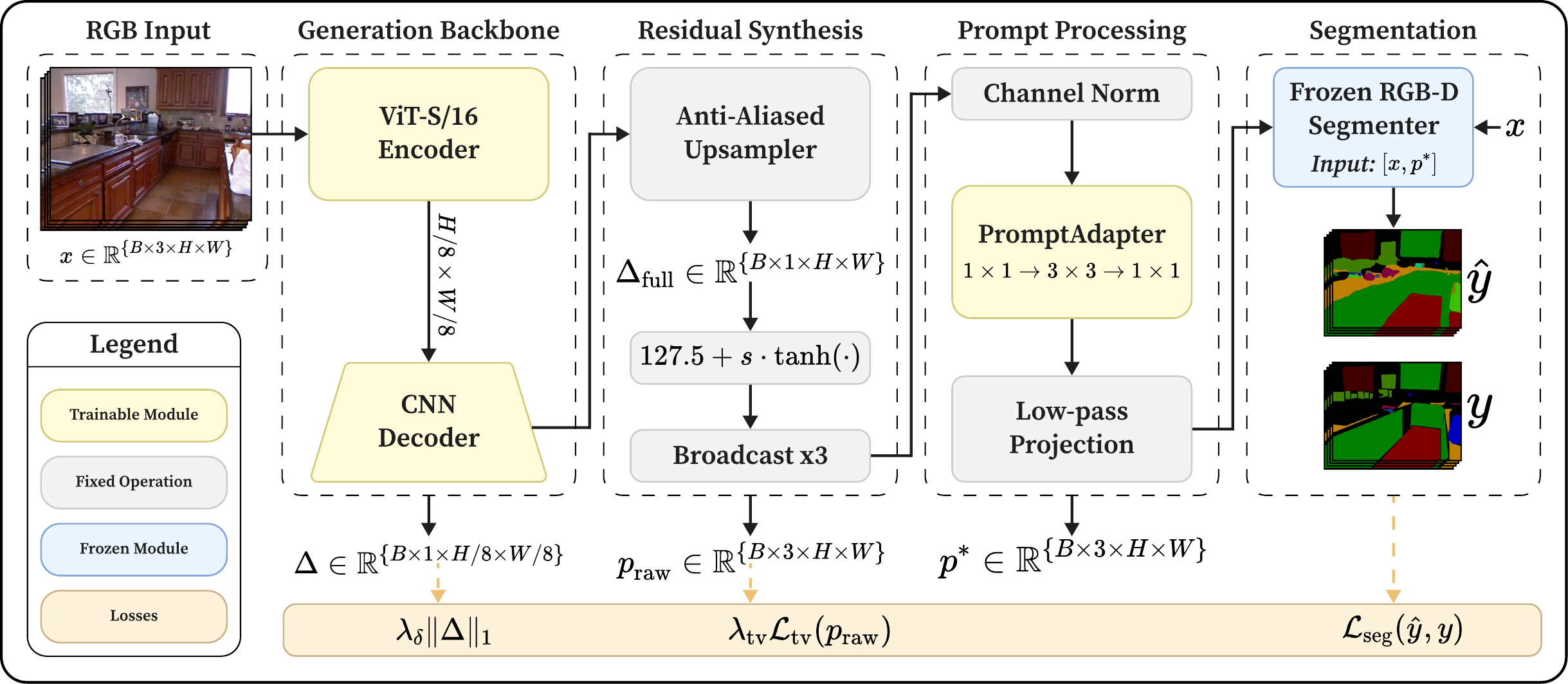}
    \caption{Overview of the GeomPrompt architecture.}
    \label{fig:gp-arch}
  \end{subfigure}
  \hfill
  \begin{subfigure}{0.375384615\linewidth}
    \includegraphics[width=\linewidth]{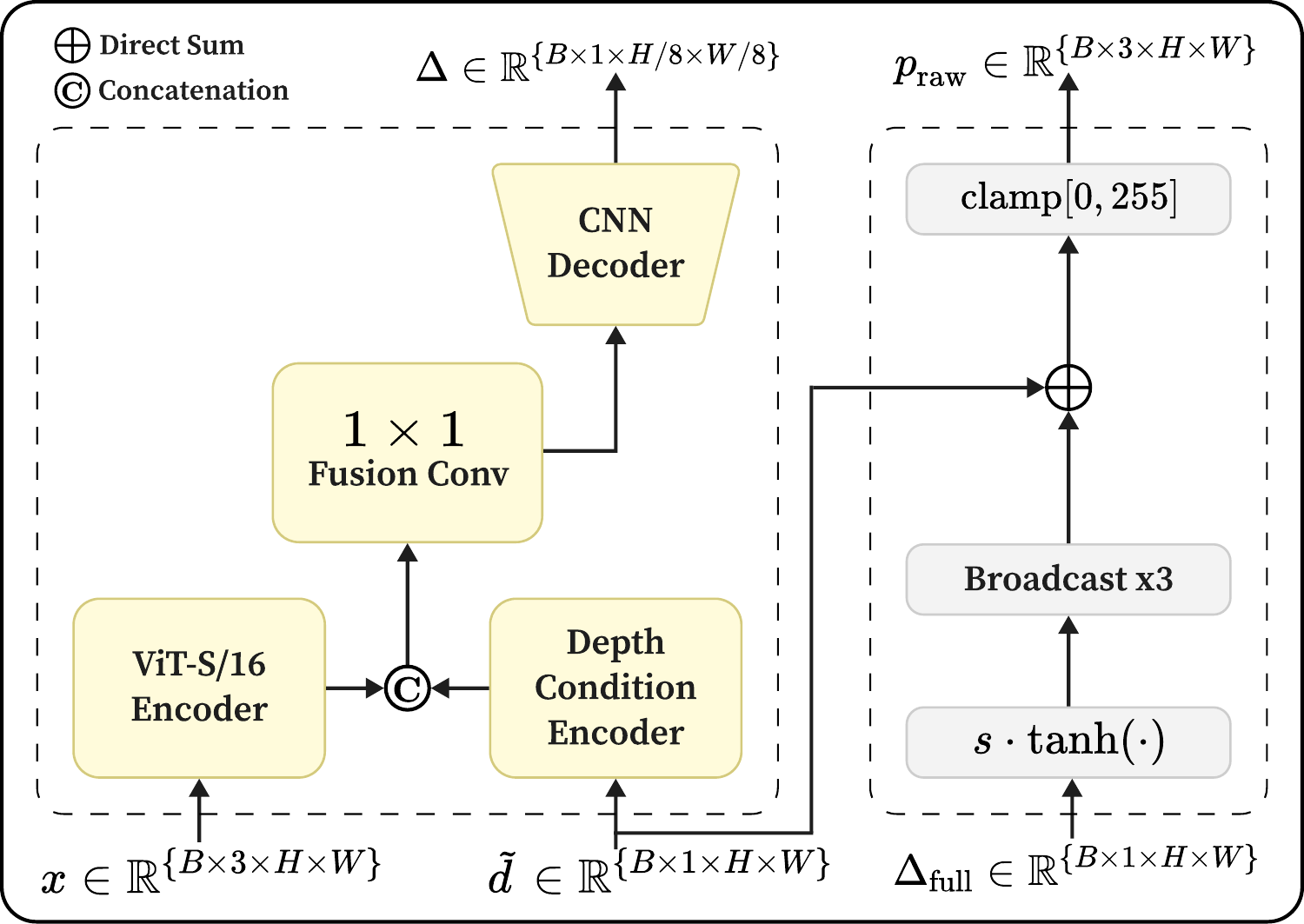}
    \caption{Depth handling in GeomPrompt-Recovery.}
    \label{fig:gpr-arch}
  \end{subfigure}
  \caption{Model architecture diagrams. (a) Overview of the GeomPrompt architecture.  (b) Depth handling in GeomPrompt-Recovery.}
  \label{fig:arch}
\end{figure*}

\subsection{GeomPrompt Architecture}
GeomPrompt accepts an ImageNet-normalized RGB image~\cite{deng_imagenet:_2009} $x \in \mathbb{R}^{3 \times H \times W}$ as input. The primary output of the module is the normalized prompt $p^*$. Auxiliary outputs generated during the forward pass include a low-resolution residual map $\Delta$ and the raw continuous prompt $p_{\text{raw}}$.

To extract rich visual semantics, the input image is processed by a ViT-S/16 encoder~\cite{dosovitskiy_image_2020}. Following the standard vision transformer forward pass, prefix tokens (such as the class token) are discarded~\cite{ranftl_vision_2021}. The remaining patch tokens are then reshaped back into a 2D spatial feature grid, preserving the spatial layout of the original image for dense prediction~\cite{ranftl_vision_2021}.

The spatial feature grid is passed to a lightweight CNN decoder consisting of an initial $\times 2$ bilinear upsampling stage followed by two $3\times3$ Conv-BN-ReLU blocks and a final $1\times1$ projection to a single-channel residual map. This yields a compact decoder that maps reassembled ViT features to a low-resolution geometric residual~\cite{ranftl_vision_2021}. Rather than predicting a full-resolution depth map directly, the decoder predicts a low-resolution residual $\Delta$ at a spatial scale of $H/8 \times W/8$. This residual is then progressively upsampled to the full $H \times W$ resolution using a fixed anti-aliased upsampler, yielding $\Delta_\text{full}$. By combining a low-resolution decoder with anti-aliased upsampling, the network suppresses spectral and checkerboard artifacts and favors more stable and smoother geometric structures over noisy high frequency per-pixel details~\cite{odena_deconvolution_2016,agnihotri_improving_2025,rahaman_spectral_2018}.

To ensure the synthesized prompt remains stable and conforms to the expected input space of the frozen segmenter, the upsampled residual undergoes a strict parameterization process. The residual is bounded using a scaled hyperbolic tangent function, $s \tanh(\cdot)$, and centered around a neutral gray prior of $127.5$. The resulting single-channel map is expanded to three channels, normalized, refined by a residual PromptAdapter, and finally passed through a hard low-pass projection before entering the segmenter. The PromptAdapter is a lightweight convolutional residual module operating in 3-channel prompt space, with a $1\times1\;\;3\to16$ projection, a $3\times3\;\;16\to16$ convolution, and a final $1\times1\;\; 16\to3$ projection. Its last layer is zero-initialized so that the module starts as an identity mapping, allowing it to learn only segmenter-specific corrections to the normalized prompt. The low-pass projection applies average-pooling downsampling followed by bilinear upsampling with a factor of 2. This suppresses high-frequency prompt artifacts and enforces a smoother geometric signal~\cite{zhang_making_2019}. This parameterization can be formalized as:
\begin{equation}
\begin{aligned}
\Delta_{\text{full}} &= \mathcal{U}(\Delta)\\
p_\text{raw} &= 127.5 + s \tanh(\Delta_{\text{full}})\\
p^* &= \Pi(\mathcal{A}(\mathcal{N}(p_\text{raw})))
\end{aligned}
\end{equation}
Where $\mathcal{U}$ denotes the fixed anti-aliased upsampler, $\mathcal{N}$ represents the normalization, $\mathcal{A}$ is the PromptAdapter, and $\Pi$ is the hard low-pass projection. We replicate the prompt to three channels before applying these operations, yielding a uniform interface for all frozen segmenters. Segmenters that consume three channel geometric input use the prompt directly, while those expecting one channel apply their standard segmenter-specific channel reduction during preprocessing. Figure~\ref{fig:gp-arch} details the end-to-end architecture of GeomPrompt.

\subsection{GeomPrompt-Recovery Architecture}

The architecture of GeomPrompt-Recovery is structured as a dual-pathway system that fuses multimodal features before predicting a structural correction. It retains the same primary RGB ViT branch utilized in the base GeomPrompt module. To process the degraded depth, we introduce an additional lightweight depth-condition encoder, where the degraded depth is first replicated to 3 channels and passed through a lightweight 4-layer $3\times3$ stride-2 CNN ($3\rightarrow32\rightarrow64\rightarrow64\rightarrow64$), producing a local structural feature map aligned to the $H/16 \times W/16$ token grid of the RGB ViT branch. The extracted RGB and depth features are concatenated along the channel dimension and subsequently fused using a $1 \times 1$ convolutional layer. This fused representation is then passed through the identical decoder and anti-aliased upsampler stack used in GeomPrompt to predict a residual correction map.

GeomPrompt-Recovery predicts a bounded residual correction that is applied directly to the incoming corrupted depth, $\tilde{d}$, in its raw continuous space. This additive correction process is formulated as:
\begin{equation}
\begin{aligned}
    \text{corr} &= s\tanh(\Delta_{\text{full}})\\
    p_{\text{raw}} &= \mathrm{clamp}(\tilde{d} + \text{corr}, 0, 255)
\end{aligned}
\end{equation}
Where $\Delta_{\text{full}}$ is the progressively upsampled output from the decoder and $s$ dictates the bounds of the scaled hyperbolic tangent function. Following this residual correction and clamping, the recovered prompt $p_{\text{raw}}$ undergoes the same downstream normalization, PromptAdapter refinement, and hard low-pass projection as defined in the base GeomPrompt module to produce $p^*$.

For principled and stable optimization, the correction head of the decoder is zero-initialized, so the model begins its training regime as an approximate identity mapping with respect to the supplied broken depth. The network is thus forced to learn purposeful deviations from the corrupted depth only when those deviations are supported by the downstream segmentation loss.

GeomPrompt-Recovery thus acts as a corruption-aware repair mechanism, learning to correct noisy or missing depth information only insofar as that recovery directly improves the semantic parsing capabilities of the frozen RGB-D segmenter. Figure~\ref{fig:gpr-arch} details the degraded depth handling that is present in GeomPrompt-Recovery.

\subsection{Training Objective and Protocol}

Across both variants of our method, the downstream RGB-D segmentation backbone remains entirely frozen; only the parameters of the prompt generation module are optimized. The primary driving signal is the segmentation loss on the final semantic prediction. To encourage stable and well-behaved prompt generation, we augment this primary objective with two regularizers: a Total Variation (TV) smoothness penalty applied to the raw continuous prompt, and an L1 magnitude penalty applied to the low-resolution residual. The overall training objective is formulated as:
\begin{equation}
    \mathcal{L} = \mathcal{L}_{\text{seg}}(\hat{y}, y) + \lambda_{\text{tv}} \mathcal{L}_{\text{tv}}(p_{\text{raw}}) + \lambda_{\delta}\|\Delta\|_1
\end{equation}
Here, $\mathcal{L}_{\text{seg}}$ represents the Online Hard Example Mining (OHEM) Cross-Entropy loss, and $\lambda_{\text{tv}}$ and $\lambda_{\delta}$ are empirical weighting coefficients.

To maintain stability during training, the residual scaling factor $s$, which bounds the prompt generation, is progressively ramped over training epochs. This allows the model to smoothly transition from predicting a safe, uniform gray prior to generating structurally complex, learned geometric features.

To equip the GeomPrompt-Recovery module with deployment robustness, we train it using a synthesized distribution of degraded depth inputs. We stochastically apply a diverse suite of spatial and systemic corruptions to GT depth, with severities dynamically varied across training iterations to expose the network to a broad spectrum of potential sensor failures.

\section{Experiments}
\label{sec:experiments}

\subsection{Implementation Details}
We train GeomPrompt and GeomPrompt-Recovery for 300 epochs using AdamW with a poly learning-rate schedule (power 0.9), 10 warmup epochs, weight decay 0.01, and separate learning rates of $3\cdot10^{-5}$ for the encoder and $1\cdot10^{-4}$ for the decoder. Training uses an effective batch size of 32 on $8\times$ A40 GPUs. We set $\lambda_{tv}=10^{-5}$ and $\lambda_{\delta}=5\times10^{-4}$, and linearly ramp the residual bound $s$ from 15 to 80 over all epochs. All training images are processed with random resize and random crop to $480\times480$. For GeomPrompt-Recovery, each training image is either kept clean with probability 0.2 or assigned exactly one uniformly sampled corruption from \{\textit{quantize, hole, dropout, noise, blur, banding, scale shift}\}. Corruption severity is sampled uniformly from $[0.10, 0.90]$. GeomPrompt outputs $p_\text{raw}\in[0,255]$, which we normalize using the DFormer~\cite{yin_dformer:_2023} depth statistics, namely $\mu = [0.48, 0.48, 0.48]$ and  $\sigma = [0.28, 0.28, 0.28]$ as $\mathcal{N}(p_\text{raw})$ for consistency across backbones.

\subsection{Experimental Setup}
We evaluate our approach on the standard test split of the SUN RGB-D dataset~\cite{song_sun_2015}. We employ DFormer~\cite{yin_dformer:_2023} and GeminiFusion~\cite{jia_geminifusion:_2024} as our base pretrained RGB-D segmentation models. These two architectures serve as complementary testbeds because they differ both in how they incorporate geometry and in how they are evaluated, as DFormer uses a dedicated hierarchical RGB-D encoder with four-stage multi-scale features and geometry-aware RGB-D blocks that combine global-awareness and local-enhancement attention, whereas GeminiFusion uses lightweight pixel-wise fusion over aligned cross-modal features within a shared four-stage transformer backbone~\cite{yin_dformer:_2023,jia_geminifusion:_2024}.

Additionally, DFormer is evaluated with multi-scale flip inference, while GeminiFusion's evaluation is reported without multi-scale or flip test-time augmentation, allowing us to test GeomPrompt under both architectural and evaluation protocol variation~\cite{yin_dformer:_2023,jia_geminifusion:_2024}.

To rigorously evaluate GeomPrompt, we compare its synthesized prompts against ground truth depth maps from SUN RGB-D as an upper bound, along with monocular depth estimators and an RGB-only baseline. 

For our depth estimators, we use state-of-the-art zero-shot monocular depth models, specifically Metric3Dv2 (ViT-small)~\cite{hu_metric3d_2024} and the standard and Hypersim~\cite{roberts_hypersim:_2021} checkpoints of Depth Anything 2 (DA2) (ViT-base)~\cite{yang_depth_2024}. We evaluate on both these checkpoints as the Hypersim checkpoint yields metric depth whereas the standard checkpoint yields relative depth. For our RGB-only baseline, we set all depth pixels to zero. This serves as a control for a complete lack of geometric signal.

Our primary evaluation metric is mean Intersection over Union (mIoU), supported by pixel accuracy (PA) as a secondary metric. To ensure a fair comparison, we adhere to a strict shared evaluation principle, as for a given model family, the RGB preprocessing, model backbone, and inference configuration, such as test-time augmentation and scaling, remain identical. The only variable altered during evaluation is the source of the depth channel input.

\subsection{RGB Geometric Prompting}
For the evaluation on DFormer, inference is conducted using standard multi-scale and flip test-time augmentation at scales of $\{0.5, 0.75, 1.0, 1.25, 1.5\}$, as is done in the original paper~\cite{yin_dformer:_2023}. We apply this on the whole image at original size, which reproduces the reported mIoU for GT depth. We use the DFormer-Base variant of DFormer. The GeminiFusion evaluation protocol uses a single-scale, no-flip inference pipeline without test-time augmentation, with images resized to $480\times480$, following the original paper~\cite{jia_geminifusion:_2024}. We use the MiT-B3 variant of GeminiFusion~\cite{xie_segformer:_2021}. Table~\ref{tab:gp_combined_eval} details our evaluation on DFormer and GeminiFusion.

\begin{table}[t]
\centering
\small
\caption{Evaluation on DFormer and GeminiFusion. GT Depth is shown as an upper bound. Best non-upper-bound results for each benchmark are in bold.}
\label{tab:gp_combined_eval}
\begin{tabular}{lcccc}
\toprule
& \multicolumn{2}{c}{\textbf{DFormer}} & \multicolumn{2}{c}{\textbf{GeminiFusion}} \\
\cmidrule(lr){2-3} \cmidrule(lr){4-5}
\textbf{Method} & \textbf{mIoU} $\uparrow$ & \textbf{PA} $\uparrow$ & \textbf{mIoU} $\uparrow$ & \textbf{PA} $\uparrow$ \\
\midrule
GT Depth              & 51.2          & 83.4          & 52.7          & 82.8          \\
\midrule
RGB-only              & 41.7          & 78.3          & 43.4          & 78.8          \\
DA2                   & 44.0          & 80.8          & \textbf{47.7} & \textbf{81.4} \\
DA2 [Hypersim]        & 47.5          & \textbf{81.8} & 44.5          & 79.6          \\
Metric3Dv2            & 46.6          & \textbf{81.8} & 46.6          & 80.6          \\
$\bigstar$ GeomPrompt & \textbf{47.8} & 81.6          & 46.4          & 80.3          \\
\bottomrule
\end{tabular}
\end{table}

On DFormer, our GeomPrompt module achieves an mIoU of 47.8 and a PA of 81.6,  outperforming by +6.1 mIoU the RGB-only baseline, which yields an mIoU of 41.7. Furthermore, GeomPrompt performs competitively with state-of-the-art explicit monocular depth estimators, as it surpasses Depth Anything 2 (44.0 mIoU), Metric3Dv2 (46.6 mIoU), and the DA2 Hypersim checkpoint (47.5 mIoU).

When evaluated on the GeminiFusion backbone, GeomPrompt attains an mIoU of 46.4 and a PA of 80.3, providing an improvement of +3.0 mIoU over the RGB-only control baseline (43.4 mIoU). The learned geometric prompt performs comparably to the Metric3Dv2 estimator (46.6 mIoU) and comfortably outperforms the DA2 Hypersim checkpoint (44.5 mIoU), although the standard DA2 checkpoint achieves the highest zero-shot depth performance in this specific setup with an mIoU of 47.7.

This demonstrates that GeomPrompt successfully learns a task-relevant geometric representation strictly from segmentation supervision, and is competitive with explicitly trained monocular depth models across different multimodal fusion paradigms.

For context, we also compare to prior missing modality methods that train the segmentation model. On SUN RGB-D, M3L reports 41.31 mIoU for RGB inference with missing depth~\cite{maheshwari_missing_2024}, and OS-MD reports 43.64 mIoU under incomplete modality evaluation~\cite{wei_one-stage_2023}. GeomPrompt achieves 47.8 mIoU with DFormer and 46.4 mIoU with GeminiFusion under missing depth, despite training only a lightweight prompt module rather than the segmenter.

\subsection{GeomPrompt-Recovery Under Depth Failures}

To evaluate the robustness of GeomPrompt-Recovery (GPR), we subjected the model to a comprehensive suite of simulated depth degradations. The full corruption family includes quantize, hole, dropout, noise, blur, banding, and scale shifting. For this detailed analysis, we focus on three primary corruptions that commonly afflict real-world depth sensors~\cite{haider_what_2022, he_depth_2017, duarte_selfredepth:_2024}.

The first degradation is quantization, which simulates limited sensor precision by mapping the depth values to discrete bins over the $[0, 255]$ range. The number of bins is bounded and inversely proportional to the severity so that higher severities yield significantly coarser depth levels~\cite{choo_statistical_2014,rached_structscan3d_2025}.

The second degradation is dropout, which mimics missing depth readings caused by reflective surfaces or sensor range limits by independently setting pixels to zero. The probability of a pixel being zeroed out is proportional to the severity. Unmasked pixels remain unchanged~\cite{costanzino_learning_2023,liu_transparent_2024,zhu_rgb-d_2021}.

Our third degradation is noise, which simulates general sensor inaccuracy via additive zero-mean Gaussian noise. The noise is applied pixelwise with a standard deviation proportional to the severity, causing saturation at the 0 and 255 boundaries at higher severities~\cite{afzal_maken_improving_2025,haider_what_2022,he_depth_2017}.

Corruptions are applied globally across the depth maps. The clean depth is first converted to a uint8 format. The specific corruption is then executed in float32 precision, with the severity parameter strictly clamped to the $[0, 1]$ range. The final output is clipped and cast back to uint8.

\begin{figure*}[t]
  \centering
  \includegraphics[width=\textwidth]{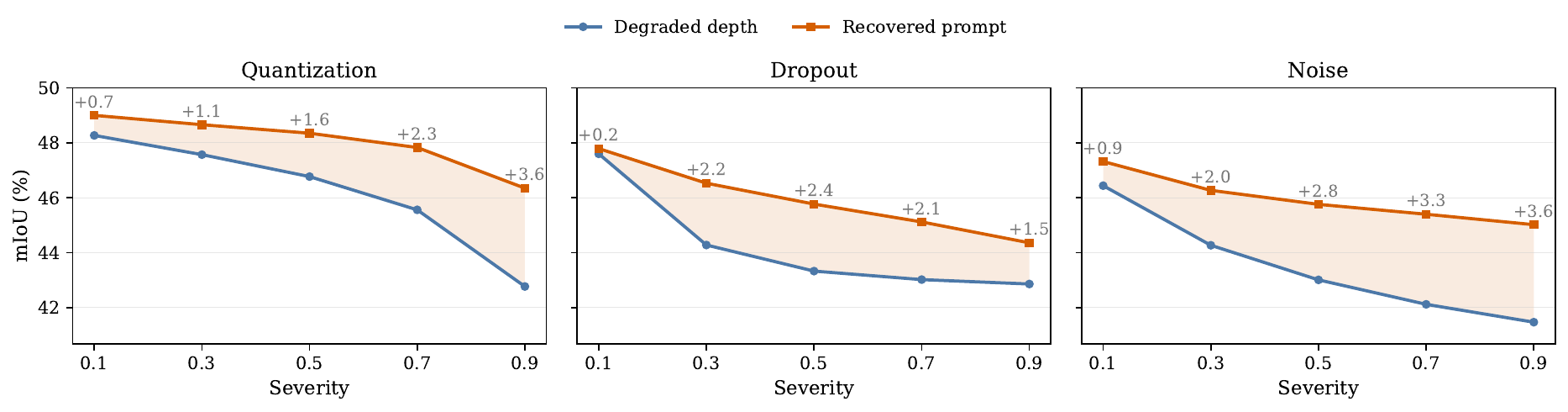}
  \caption{Degraded depth vs. our recovered prompt across severities and degradation types.}
  \label{fig:gpr_plot}
\end{figure*}

Both the degraded depth and the GPR recovered prompt are passed into DFormer to evaluate segmentation results. Figure~\ref{fig:gpr_plot} illustrates the mIoU performance of the baseline degraded depth versus the recovered prompt across severities ranging from 0.1 to 0.9. Across all evaluated corruptions, GPR consistently acts as a buffer against catastrophic failure, maintaining higher mIoU than the raw degraded depth. Table~\ref{tab:degradation_gains} summarizes the mIoU improvements yielded by the recovered prompt compared to the degraded depth baseline.

\begin{table}[t]
\centering
\small
\caption{Recovered gains across different degradation types.}
\label{tab:degradation_gains}
\begin{tabular}{lcc}
\toprule
\textbf{Degradation} & \textbf{Mean mIoU Gain} $\uparrow$ & \textbf{High-Severity Gain} $\uparrow$ \\
\midrule
Quantization & +1.4 & +2.3 \\
Dropout  & +2.0 & +1.5 \\
Noise    & +2.5 & +3.6 \\
\bottomrule
\end{tabular}
\end{table}

The results indicate that GPR effectively recovers task-relevant geometric cues even when the input depth is severely compromised. Notably, the performance gap between the recovered prompt and the degraded depth generally widens as the severity increases. For example, while high-severity noise causes severe pixel saturation, GPR is still able to recover up to +3.6 mIoU.

This validates our core hypothesis that by predicting a bounded residual correction from the RGB input, GPR can partially restore the structured prior that is useful for the frozen segmenter. Because it is trained purely under a segmentation objective, GPR bypasses the need for explicit metric depth reconstruction, allowing it to dynamically compensate for structural failures in the depth map using contextual visual cues.

\subsection{Analysis}
\subsubsection{Inference Cost}

\begin{table}[t]
\centering
\small
\setlength{\tabcolsep}{4pt} 
\caption{Efficiency comparison across methods. Lower is better.}
\label{tab:efficiency_comparison}
\begin{tabular}{lccc}
\toprule
\textbf{Model} & \textbf{Lat. (ms)} $\downarrow$ & \textbf{FLOPs (G)} $\downarrow$ & \textbf{Params (M)} $\downarrow$ \\
\midrule
DA2            & 38.3         & 280.7         & 97.5 \\
Metric3Dv2     & 71.9         & 315.2         & 37.5 \\
$\bigstar$ GeomPrompt & \textbf{7.8} & \textbf{44.0} & \textbf{23.4} \\
\bottomrule
\end{tabular}
\end{table}

Since standard metric depth estimators are often computationally heavy~\cite{schiavella_efficient_2025, zhang_survey_2025}, we evaluate the efficiency of GeomPrompt against Depth Anything 2 (ViT-B) and Metric3Dv2 (ViT-S). The evaluation uses a batch size of 1 on a single A40 GPU with fp16 autocast precision. Latency includes only model execution time. Parameter counts reflect total inference-time parameters, and FLOPs are computed per forward pass on the respective preprocessed tensors. Table ~\ref{tab:efficiency_comparison} details our efficiency comparison results.

GeomPrompt is substantially lighter than monocular depth baselines, as at 7.8 ms per frame, it is nearly 5 times faster than DA2 (38.3 ms) and over 9 times faster than Metric3Dv2 (71.9 ms). Furthermore, GeomPrompt requires only 44.0 GFLOPs and 23.4 million parameters compared to DA2's 280.7 GFLOPs and 97.5 million parameters. This lightweight profile suggests that GeomPrompt can be practical as an inline plugin for real-time embodied agents in environments with unreliable sensor depth.

\subsubsection{Naive Depth Channel Controls}

To verify GeomPrompt learns genuine geometric representations rather than exploiting low-level statistics, we evaluate it against four handcrafted pseudo-depth baselines. Luminance Gray (grayscale intensity) tests if brightness replaces geometry~\cite{barron_shape_2012, geirhos_shortcut_2020}. Canny Distance Zero Blend blends a normalized Canny edge distance transform with zero depth to test edge-distance cues~\cite{canny_computational_1986,felzenszwalb_distance_2012}.  Laplacian Edges uses an absolute Laplacian response normalized to $[0, 255]$ as a second-order structural prior~\cite{marr_theory_1980}. Lastly, Scharr Distance Zero Blend applies an L2 distance transform to thresholded Scharr gradients, followed by normalization, Gaussian smoothing, and zero-depth blending for a smooth boundary prior~\cite{scharr_optimal_2007, felzenszwalb_distance_2012}. Table~\ref{tab:naive_baselines_miou} details our results against these baselines.

\begin{table}[t]
\centering
\small
\caption{Naive baselines and references, separated by segmenter. Best results in each segmenter group are in bold.}
\label{tab:naive_baselines_miou}
\begin{tabular}{lcc}
\toprule
\textbf{Method} & \textbf{DFormer} $\uparrow$ & \textbf{GeminiFusion} $\uparrow$ \\
\midrule
RGB-only                     & 41.7          & 43.4          \\
Luminance Gray               & 25.4          & 43.2          \\
Laplacian Edges              & 40.3          & 42.4          \\
Scharr                       & 39.8          & 43.0          \\
Canny                        & 38.6          & 43.5          \\
$\bigstar$ GeomPrompt & \textbf{47.8} & \textbf{46.4} \\
\bottomrule
\end{tabular}
\end{table}

Handcrafted pseudo-depth proxies consistently fail to surpass the RGB-only baseline across both backbones. On DFormer, naive controls degrade the 41.7 mIoU RGB baseline, dropping to 25.4 mIoU for luminance and peaking at 40.3 mIoU for Laplacian edges. Similarly, GeminiFusion's 43.4 mIoU baseline sees naive controls plateau between 42.4 and 43.5 mIoU. Conversely, GeomPrompt achieves 47.8 mIoU (DFormer) and 46.4 mIoU (GeminiFusion). The failure of classical structural cues confirms GeomPrompt extracts task-relevant geometric embeddings rather than merely highlighting edges or intensity gradients.

\subsubsection{Training Ablations}

\begin{figure}[t]
  \centering
  \includegraphics[width=.95\columnwidth]{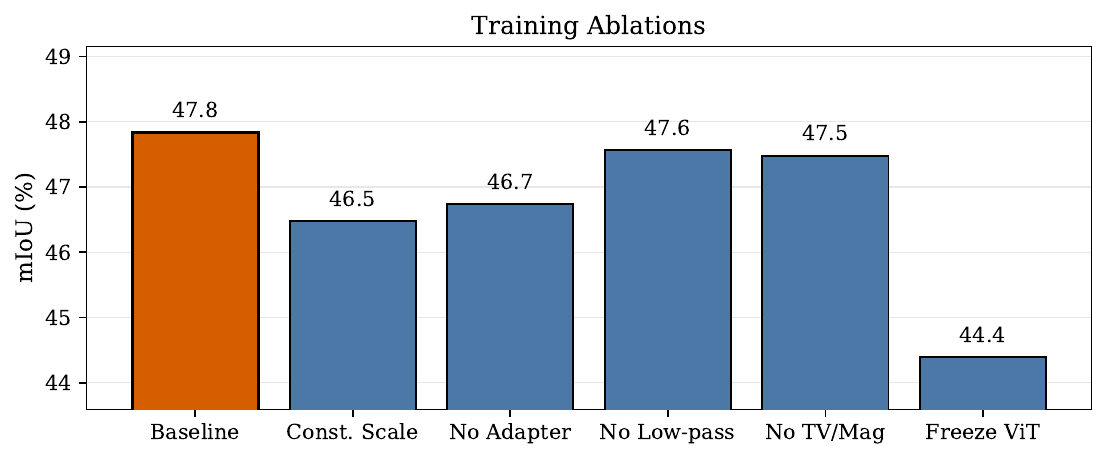}
  \caption{mIoU of training ablations vs. our baseline.}
  \label{fig:ablations}
\end{figure}

We ablate key architectural and training components of GeomPrompt to isolate their contributions to the final performance on the SUN RGB-D test set. Figure~\ref{fig:ablations} shows the mIoU of all ablations with respect to the baseline.

A constant residual scale ($s=80$) is maintained, which isolates the value of the residual-scale curriculum by removing it and starting at maximum correction strength from epoch 1. The 1.4 mIoU drop suggests that linearly scaling the residual contribution during training acts as a stabilizer for prompt construction.

The PromptAdapter is disabled, which causes a modest but noticeable degradation, showing its utility in refining the predicted prompt to better match the expected distribution of the frozen segmenter.

The low-pass projection is removed, which tests whether suppressing high-frequency residual structure is important for effective prompt construction. The resulting performance drops only slightly by 0.2 mIoU, indicating that the low-pass projection is not a primary driver of the gains.

The TV and magnitude regularization are disabled, which tests whether explicit smoothness and amplitude control are necessary for learning useful prompts. This causes a minor 0.3 mIoU degradation, suggesting that GeomPrompt does not rely heavily on them to achieve its improvement.

The ViT encoder is frozen, which tests whether end-to-end ViT feature adaptation is necessary. This results in a severe performance drop of 3.5 mIoU compared to the baseline, confirming that the base visual features must adapt to effectively synthesize geometric prompts.

\subsection{Qualitative Results}
We qualitatively evaluate our prompts to illustrate how they adapt to the absence or degradation of metric depth.

\begin{figure}[t]
  \centering
  \includegraphics[width=\columnwidth]{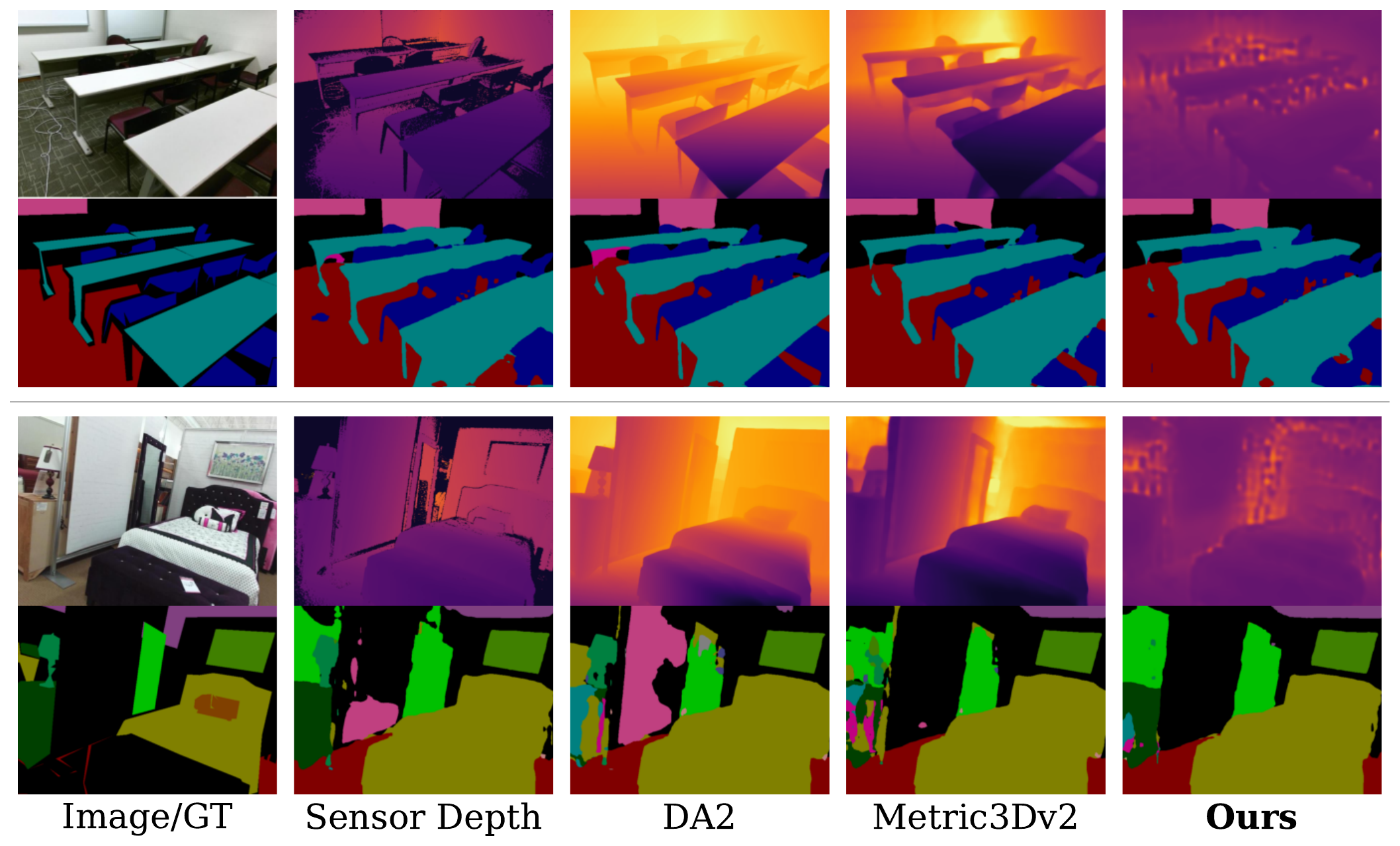}
  \caption{Qualitative comparison of segmentation outputs and geometric representations.}
  \label{fig:qual_segmentation}
\end{figure}

\begin{figure}[t]
  \centering
  \includegraphics[width=.7\columnwidth]{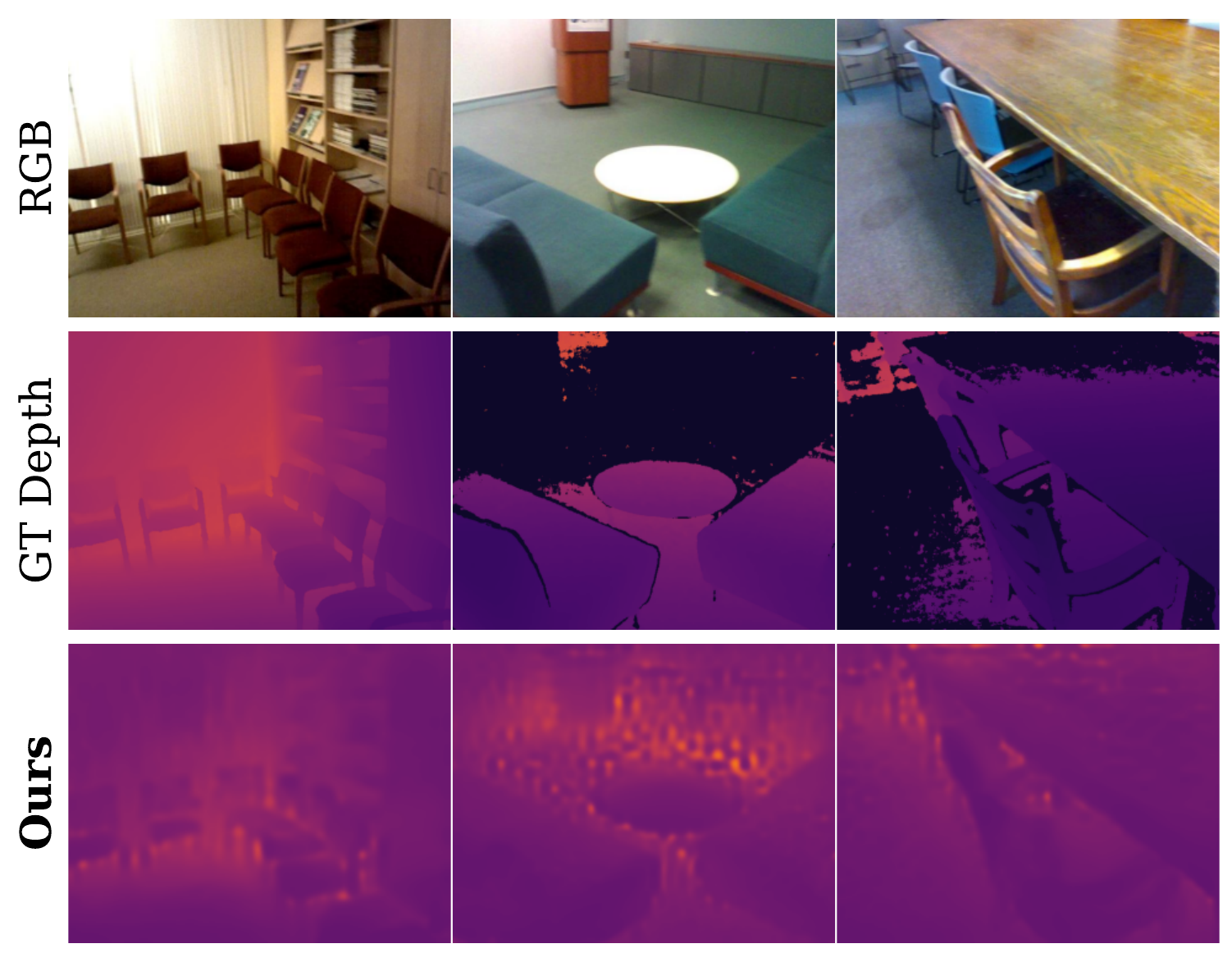}
  \caption{Visual comparison of our synthesized geometric prompt against the original RGB input and ground truth depth.}
  \label{fig:qual_prompt_vs_depth}
\end{figure}

Figure~\ref{fig:qual_segmentation} shows that GeomPrompt produces segmentation outputs competitive with depth-based alternatives while generating prompts that differ visibly from metric depth, emphasizing boundaries, planar structure, and semantic grouping. Figure~\ref{fig:qual_prompt_vs_depth} highlights this distinction by comparing our prompt and GT depth across images. 

\begin{figure}[t]
  \centering
  \includegraphics[width=.95\columnwidth]{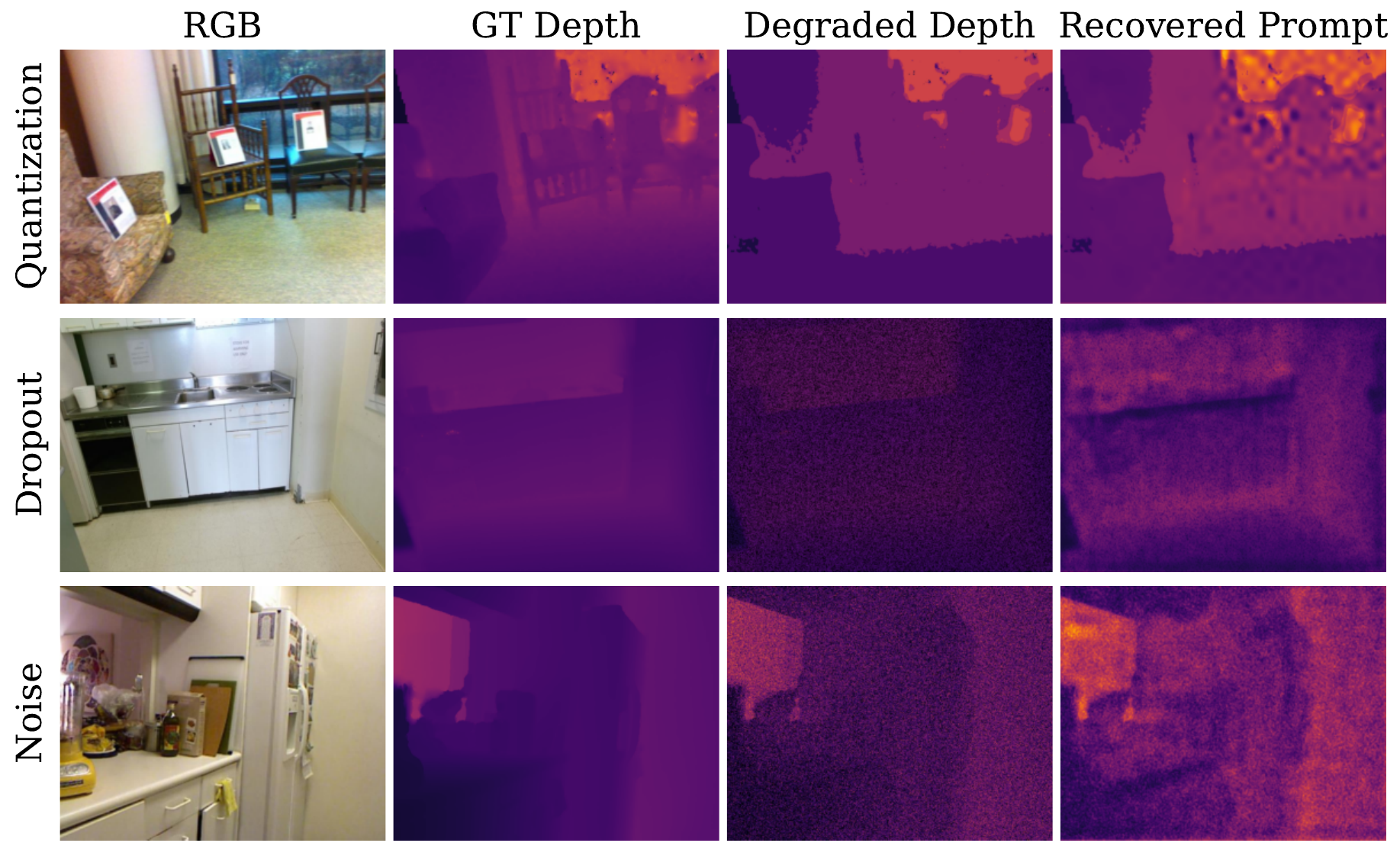}
  \caption{GeomPrompt-Recovery under simulated depth failures.}
  \label{fig:qual_degradation}
\end{figure}

Figure~\ref{fig:qual_degradation} demonstrates the efficacy of GeomPrompt-Recovery (GPR) when subjected to severe sensor failures. GPR successfully processes the corrupted data to output a recovered representation that effectively smooths out high-frequency artifacts and appears to restore object boundaries.

\section{Conclusion}
\label{sec:conclusion}
In this work, we presented GeomPrompt and GeomPrompt-Recovery as lightweight cross-modal adaptation modules for RGB-D perception under missing and degraded depth. Instead of reconstructing metric depth, the method learns task-driven geometric prompts from downstream supervision alone, enabling a frozen multimodal segmenter to remain effective when one sensing stream is unavailable or corrupted. GeomPrompt improves over RGB-only inference while remaining competitive with monocular depth estimators, and GeomPrompt-Recovery improves robustness under simulated sensor failures. These results suggest that task-driven cross-modal compensation can be a practical strategy for robust multimodal segmentation in embodied systems, especially when real-world deployment requires efficiency and graceful degradation under unreliable sensing. Future work could study whether this prompting view extends to other multimodal perception tasks useful for embodied AI, such as mapping, navigation, or manipulation.

{
    \small
    \bibliographystyle{ieeenat_fullname}
    \bibliography{main}
}

\end{document}